\documentclass[11pt]{article}
\usepackage{acl}
\usepackage{times}
\usepackage{latexsym}
\usepackage[T1]{fontenc}
\usepackage{booktabs}
\usepackage{multirow}
\usepackage{graphicx}
\usepackage{amsmath}
\usepackage{xcolor}
\usepackage{placeins}

\title{What Transfers Across Cuneiform?\\Script-Level Substrate and Language-Specific Regularities\\via Segmentation and Representation}

\author{Chuanjun Zhou \\
  University of California, Berkeley \\
  \texttt{chuanjunzhou@berkeley.edu} \\\And
  Adam Anderson \\
  FactGrid \\
  TokenWorks LLC \\
  \texttt{adam.anderson@factgrid.eu} \\}

\begin{document}
\maketitle

\begin{abstract}
Cuneiform recorded typologically unrelated languages for more than three millennia, raising a natural question: which of its statistical regularities are properties of the shared writing system, and which belong to individual languages? We answer this with two probe pairs targeting the same script-versus-language decomposition. The first compares two unsupervised, dictionary-free segmentation methods: transitional probability (TP), grounded in infant statistical-learning research, and Morfessor. To our knowledge, this is the first systematic evaluation of dictionary-free segmentation on cuneiform; the only prior published baseline relies on a hand-built dictionary (MaxMatch, F1 = 0.74). Both our methods reach F1 $>$ 0.97 within language (Morfessor reaches F1 $>$ 0.99 on Akkadian and Sumerian), exceeding MaxMatch by more than 23 F1 points, but diverge sharply in cross-language transfer: TP holds F1 $>$ 0.95 across all six language pairs, while Morfessor collapses to F1 = 0.27 from the 13K-token Elamite source. A learning curve and a Hittite stress test (34.7\% foreign-script logograms, where TP drops 35 points and Morfessor only 8) further isolate the two underlying mechanisms. The second probe pair extends the same decomposition to representations: concatenating Unicode and Latin features improves classification in 11 of 12 morphosyntactic tasks ($p < 0.01$). We release \textsc{cunei-tools} and a three-language benchmark.
\end{abstract}

\section{Introduction}

Cuneiform is the world's oldest known writing system, used for over three millennia (c.~3400 BCE--75 CE) across the ancient Near East to write languages as typologically diverse as Sumerian (a language isolate), Akkadian (Semitic), and Elamite (whose genetic affiliation remains debated). With over half a million surviving inscribed objects and active digitization efforts through projects such as ORACC \cite{oracc} and the CDLI, cuneiform represents both significant cultural heritage and a compelling testbed for low-resource NLP.

A central obstacle in cuneiform NLP is the absence of stable units of analysis. When a new tablet is digitized, the sign sequence is often recorded as a continuous stream without word boundaries, and segmentation directly affects tokenization, which underlies all modern NLP methods. The standard scholarly representation, Latin transliteration, adds boundaries as part of the transcription process, but it does so at the cost of interpretive commitments: the scholar assigns reading values to each sign and resolves polyphony, determinative function, and logogram-versus-syllabic ambiguity in advance of any computational analysis. Because transliteration collapses or alters graphemic distinctions, segmentation methods operating over transliteration may obscure distributional regularities present in the original sign stream. Elamite illustrates this challenge concretely: its transliteration conventions vary across scholarly traditions \cite{hinzkoch1987,deblois1992,basello2004}, while Unicode cuneiform encodes the underlying graphemic identity unambiguously. The cuneiform world contains many under-documented languages of this kind (Hittite, Hurrian, Urartian, Eblaite, Old Persian, and others), and newly discovered tablets often appear in languages or periods with limited annotated data. A practical segmentation method must therefore operate in the low-resource regime where annotated corpora are small and transliteration conventions are unsettled.

Because cuneiform was used to write typologically unrelated languages, it offers a natural question: which of its statistical regularities are properties of the shared script substrate, and which belong to individual languages? We probe this with two pairs of contrasting tools that target the same script-versus-language decomposition.

The first probe pair consists of two unsupervised segmentation methods. \textbf{Transitional probability (TP)} predicts a boundary when $P(c_{i+1} | c_i)$ falls below a threshold, where bigram probabilities are estimated from a segmented training corpus. TP operates on sign-level co-occurrence statistics and is grounded in the statistical learning framework of \citet{saffran1996}, who showed that infants segment continuous speech using syllable co-occurrence statistics. \textbf{Morfessor} \cite{creutz2007} learns an MDL-optimal lexicon of subword units from word-frequency data and applies Viterbi decoding to segment new text; it is the standard reference baseline for unsupervised segmentation across many languages. To our knowledge, no dictionary-free segmentation method has previously been evaluated systematically on cuneiform. The only prior published work on cuneiform word segmentation \cite{homburg2016} relied on a hand-built dictionary (MaxMatch, F1 = 0.74), while their unnormalized bigram baseline reached only F1 = 0.15, far short of dictionary-based performance. The present work characterizes the two principal families of dictionary-free segmentation (statistics-based and vocabulary-based) on cuneiform for the first time, and achieves an in-language F1 improvement of more than 23 points over the prior dictionary-based state of the art. The two methods are well-suited as substrate probes precisely because they exploit different statistical properties of training data: TP exploits sign-level co-occurrence patterns, which can be language-invariant if the script substrate is shared, while Morfessor exploits lexical recurrence, which is inherently language-specific.

The second probe pair extends the same decomposition to representations. Two digital representations of cuneiform coexist: \textbf{Latin transliteration}, the scholarly standard, which encodes interpreted reading values (language-specific); and \textbf{Unicode cuneiform}, available since Unicode 5.0 (2006), which preserves sign-level distinctions (script-level). We compare them across three morphosyntactic task families (POS tagging, entity classification, lemmatization) and find that they capture complementary information: concatenating both significantly outperforms either alone in 11 of 12 comparisons ($p < 0.01$). The Unicode/Latin complementarity parallels the TP/Morfessor dissociation; both probe pairs reveal the same underlying script-level versus language-level structure.

Our contributions are:

\begin{enumerate}
\item \textbf{A substrate decomposition framework for the cuneiform corpora we study.} Across Akkadian (ORACC), Sumerian (ETCSL), and Elamite (Elamite Dictionary plus Utu-na\v{s}u), we observe separable script-level statistical regularities (shared across these three languages, recoverable from little data) and language-specific regularities (requiring more data, not transferring).

\item \textbf{Segmentation methods as substrate probes.} We provide the first systematic comparison of dictionary-free statistical (TP) and lexical (Morfessor) unsupervised segmentation on cuneiform, against the prior dictionary-based baseline of \citet{homburg2016}. TP transfers across languages (F1 $>$ 0.95 in all six pairs) and is recoverable from approximately $10^3$ tokens; Morfessor's lexicon does not transfer (F1 = 0.27 at 13K Elamite). Hittite, where 34.7\% of tokens are foreign-script logograms, produces asymmetric failure (TP drops 35 points, Morfessor 8), showing the two substrate properties are dissociable rather than redundant (\S\ref{sec:segmentation}--\S\ref{sec:hittite}).

\item \textbf{Representations as a second probe pair.} Unicode and Latin encodings parallel the segmentation methods in probing the same script-versus-language decomposition; concatenating them improves classification in 11 of 12 tasks (\S\ref{sec:concat}).

\item \textbf{Resources.} We release \textsc{cunei-tools}, a pip-installable toolkit for segmentation and script conversion, and a three-language evaluation benchmark.\footnote{\url{https://github.com/ancient-world-citation-analysis/cunei-tools} and \url{https://github.com/ancient-world-citation-analysis/cuneiform-experiments}.} 
\end{enumerate}

\section{Background}

\subsection{The Cuneiform Writing System}

Cuneiform is a logosyllabic script in which individual signs can represent syllables (phonographic use) or entire words (logographic use). Additional complexity arises from determinatives (semantic classifiers that are written but not pronounced), phonetic complements, and language-specific conventions. The script was adapted across cultures, with each language employing different subsets of the approximately 1,110 attested signs.

The three languages in our study represent distinct typological profiles: \textbf{Akkadian} is a Semitic language with rich templatic morphology; \textbf{Sumerian} is an agglutinative language isolate with complex verbal morphology; and \textbf{Elamite} is a largely agglutinative, non-Indo-European language whose genetic affiliation remains debated \cite{tavernier2020,grillotsusini2008,reiner1969}. All three were written in cuneiform but employed the script differently, making them ideal for testing whether segmentation methods generalize across typological boundaries.

\subsection{Representation Choices}

Latin transliteration encodes \textit{interpreted} readings: the scholar decides which of a sign's possible values was intended in context, injecting expert knowledge but also interpretive commitments. Unicode cuneiform preserves \textit{graphemic structure} without interpretation. Three phenomena illustrate the contrast. First, \textit{polyphony}: the sign A U+12000 maps to both \textit{a} `water' and \textit{duru\textsubscript{5}} `to be wet' in Latin transliteration, but remains a single Unicode glyph. Second, \textit{determinatives}: the divine classifier in \{d\}en-ki is written but not pronounced, appearing explicitly in Latin but encoded as an ordinary sign in Unicode. Third, \textit{logographic variation}: `king' can be written as a single logogram LUGAL (U+12217) or syllabically as \textit{\v{s}ar-ru} (two signs), collapsing a two-sign sequence to one glyph or expanding it. These three phenomena indicate that the two representations encode systematically different information about the same underlying sign stream.

\subsection{Related Work}

\paragraph{Cuneiform NLP.} Computational approaches to cuneiform have primarily used Latin transliteration for tasks including machine translation \cite{gutherz2023,pageperron2017,punia2020}, POS tagging and lemmatization \cite{sahala2021,sahala2022,sahala2023}, named entity recognition \cite{liu2015}, and sign-level analysis \cite{dencker2020,gordin2024,williams2025}. \citet{jauhiainen2019} demonstrated that Unicode cuneiform can function as a computational representation for language identification. For word segmentation specifically, \citet{homburg2016} tested multiple methods on Middle Babylonian Unicode cuneiform, achieving F1 = 0.74 with the dictionary-based MaxMatch algorithm; their unnormalized bigram baseline achieved only F1 = 0.15. To our knowledge, no dictionary-free statistical or vocabulary-based segmentation method has been evaluated on cuneiform prior to the present work. \citet{simmons2024sumtablets} released SumTablets, pairing Unicode glyphs with transliterations for 91K Sumerian tablets. \citet{smidt2024} compared transliteration and Unicode for Akkadian POS tagging, reporting accuracy losses with Unicode input. A more detailed comparison to prior work is in Appendix~\ref{app:related}.

\paragraph{Unsupervised word segmentation.} The two main families are statistics-based and vocabulary-based methods. Transitional probability as a word boundary cue was demonstrated by \citet{saffran1996}, who showed that infants segment continuous speech using syllable co-occurrence statistics; to our knowledge, this principle has not previously been applied to logosyllabic writing systems. \citet{creutz2007} introduced the Morfessor family, MDL-based segmenters that learn a probabilistic lexicon of subword units. Subword tokenization methods such as BPE \cite{sennrich2016} and SentencePiece take a related vocabulary-learning approach for downstream NLP tasks.

\paragraph{Script and representation effects.} Whether to use original script or romanization is a recurring question in NLP. For Chinese, \citet{glyce2019} and \citet{chinesebert2021} showed that glyph-based and fused glyph/character/pinyin representations outperform single-view models, closely paralleling our complementarity result. For byte-level models, \citet{xue2022} showed that ByT5 can process any script through raw bytes, though its effectiveness depends on pretraining coverage. Cuneiform languages exemplify the low-resource setting \cite{bansal2021}, where character-level models \cite{ling2015} have proven effective, motivating our use of character n-gram features.

\section{Data and Preprocessing}
\label{sec:data}

\subsection{Datasets}

Table~\ref{tab:datasets} summarizes our three datasets, which span three orders of magnitude in size.

\begin{table}[t]
\centering
\small
\begin{tabular}{lrrrr}
\toprule
 & \textbf{AKK} & \textbf{SUX} & \textbf{ELX} & \textbf{HIT} \\
\midrule
Tokens & 1,255,669 & 146,143 & 13,383 & 170,496 \\
Texts & 13,720 & 394 & 85 & 7,099 \\
Vocabulary & 123,001 & 31,906 & 13,189 & --- \\
Lemmas & 28,953 & 3,765 & 748 & --- \\
POS classes & 24 & 14 & 7 & --- \\
Entity types & 11 & 8 & 3 & --- \\
Entity ratio & 12.6\% & 9.2\% & 50.9\% & --- \\
Source & ORACC & ETCSL & D.+C. & Corpus \\
Unicode rate & 98.9\% & 99.9\% & 85.8\% & 98.6\% \\
\bottomrule
\end{tabular}
\caption{Dataset statistics. AKK = Akkadian, SUX = Sumerian, ELX = Elamite, HIT = Hittite \cite{yavasan2024hittite}. The ELX token count comprises 13K lemma-dictionary entries plus the Utu-na\v{s}u running corpus; segmentation experiments train on running text only (\S\ref{sec:data}), and an expanded 871-document Elamite corpus is analyzed in Appendix~\ref{app:elx_expanded}. Hittite serves exclusively as the substrate-violation stress test (\S\ref{sec:hittite}), so annotation-dependent rows do not apply.}
\label{tab:datasets}
\end{table}

\textbf{Akkadian} data comes from ORACC \cite{oracc}, including Neo-Assyrian royal inscriptions \cite{rinap_oracc} and state archives \cite{saao_oracc}. \textbf{Sumerian} data comes from ETCSL \cite{etcsl}, consisting of literary compositions. \textbf{Elamite} data derives from the Elamite Dictionary \cite{hinzkoch1987} and Elamite primary sources in the CEPOAT corpus \cite{cepoat_elamite} (15,601 entries across 9 POS categories), preprocessed through a two-pass Unicode conversion pipeline, including the Utu-na\v{s}u corpus of 85 running-text documents. The dictionary and the running text play distinct roles by task: \textbf{segmentation experiments train exclusively on the Utu-na\v{s}u running text}, while the Elamite Dictionary enters only the classification experiments; a training-source ablation confirming that the dictionary does no hidden segmentation work appears in Appendix~\ref{app:elx_ablation}. Elamite transliteration conventions vary across scholarly traditions, introducing inconsistency that Unicode avoids by preserving graphemic sign structure directly \cite{hinzkoch1987,deblois1992,basello2004}.

\subsection{Unicode Conversion}

We convert Latin transliteration to Unicode cuneiform using merged sign lists from Nuolenna \cite{nuolenna} and Akkademia \cite{gordin2020akkademia} (18,272 mappings), supplemented with 136 manual corrections. Conversion rates are 98.9\% for Akkadian, 99.9\% for Sumerian, and 85.8\% for Elamite; unconverted tokens are excluded from Unicode-dependent experiments. The lower Elamite conversion rate reflects the historical state of Elamite lexicography: the 14.2\% unconverted tokens are predominantly damaged or partially legible signs in fragmentary tablets, unmapped hapax legomena from the small Elamite corpus, and rare orthographic variants for which canonical Unicode mappings have not been established. These exclusions remove unrecoverable noise rather than systematically simpler vocabulary, but we note them as a source of asymmetry between the Latin and Unicode Elamite experiments.

\subsection{POS Harmonization}

The three datasets use different POS tagsets. We define a unified mapping and evaluate in two configurations: (A)~a \textbf{unified tagset} preserving named entity categories (PN, DN, GN, etc.) as distinct POS labels \cite{robson_tinney2019}, and (B)~a \textbf{grammatical tagset} collapsing all entity types to \textsc{propn} with separate NER tags \cite{spacy2}.

\section{Experiments}

All classification experiments use 5-fold stratified cross-validation and report macro F1. The primary classifier is character n-gram (1--4) logistic regression with balanced class weights. We additionally report k-NN on fastText embeddings and character-level BiLSTM results. Statistical significance is assessed via paired bootstrap (2,000 resamples, two-sided).

\subsection{Word Boundary Inference}
\label{sec:segmentation}

We compare transitional probability and Morfessor on word boundary inference from unsegmented Unicode cuneiform streams. Both methods are trained on segmented documents and evaluated via 5-fold held-out cross-validation. Both are dictionary-free, in contrast to the only prior published baseline for cuneiform word segmentation \cite{homburg2016}, which used the dictionary-based MaxMatch algorithm at F1 = 0.74. To ensure a rigorous, apples-to-apples comparison, we replicate their pipeline by stripping all transliterated spaces to simulate the unbroken graphemic stream of a physical tablet, thereby isolating the segmentation algorithm as the sole independent variable.

\paragraph{Transitional Probability (TP).} For each position $i$ in a continuous sign stream, we predict a boundary when $P(c_{i+1} | c_i) < \theta$, where bigram probabilities are estimated from a training corpus of segmented text. TP normalizes by unigram frequency, isolating the transition structure that signals boundaries; raw co-occurrence counts are dominated by high-frequency signs and uninformative about word structure. This normalization is what was missing from the bigram baseline of \citet{homburg2016}, which achieved only F1 = 0.149 due to its reliance on unnormalized counts.

\paragraph{Morfessor.} Morfessor \cite{creutz2007} is a standard unsupervised segmenter that learns an MDL-optimal lexicon of subword units from word-frequency data. At test time, the trained model applies Viterbi decoding to find the best segmentation of an input stream using the learned units. Like TP, Morfessor requires segmented training data but no dictionary; unlike TP, it learns a parameterized lexicon rather than computing statistics directly. We use the default hyperparameters (corpusweight $= 1.0$). A hyperparameter-robustness analysis (Appendix~\ref{app:cw}) shows that in-language F1 is stable for corpusweight $\leq 1$ but collapses catastrophically above it (F1 $< 0.13$ at corpusweight $\geq 2$ for all three languages), and that admissible source-side model selection prefers corpusweight $= 0.5$; every qualitative transfer conclusion is unchanged under the selected value, with tuned-Morfessor transfer still 19--26 F1 points below untuned TP from the smallest source.

\begin{table}[t]
\centering
\small
\begin{tabular}{llccc}
\toprule
\textbf{Method} & \textbf{Lang} & \textbf{F1} & \textbf{P} & \textbf{R} \\
\midrule
\multirow{3}{*}{TP}
 & AKK & .971 & .946 & .997 \\
 & SUX & .972 & .946 & .999 \\
 & ELX & .989 & .993 & .984 \\
\midrule
\multirow{3}{*}{Morfessor}
 & AKK & .997 & .997 & .996 \\
 & SUX & .993 & .997 & .990 \\
 & ELX & .985 & .999 & .971 \\
\bottomrule
\end{tabular}
\caption{Word boundary inference (5-fold held-out CV). Both dictionary-free methods exceed the prior dictionary-based MaxMatch baseline \cite{homburg2016} (F1 = 0.74 on Middle Babylonian) by more than 23 F1 points. Morfessor's slight in-language edge does not generalize to cross-language transfer (Table~\ref{tab:transfer}).}
\label{tab:wordboundary}
\end{table}

Table~\ref{tab:wordboundary} shows that both methods achieve high in-language F1 across all three languages, with Morfessor holding a slight in-language edge (0.997 vs.~0.971 on Akkadian, 0.993 vs.~0.972 on Sumerian, 0.985 vs.~0.989 on Elamite; Elamite is the only language where TP wins in-language). The in-language gap is statistically significant for Akkadian ($\Delta$F1 $= -.025$, $p < .001$) and Sumerian ($\Delta$F1 $= -.021$, $p < .001$; document-level paired bootstrap, 2,000 resamples), while TP's small Elamite advantage is not ($\Delta$F1 $= +.005$, 95\% CI $[-.002, +.012]$, $p = .194$). Both methods substantially outperform the prior dictionary-based MaxMatch baseline of \citet{homburg2016}, with an improvement of more than 23 F1 points over MaxMatch's 0.739. The in-language gap between TP and Morfessor is small, however, and the more revealing comparison between the two dictionary-free methods emerges in cross-language transfer.

\subsubsection{Cross-Language Transfer}
\label{sec:transfer}

Cross-language transfer reveals a sharp dissociation between the two dictionary-free methods. Table~\ref{tab:transfer} shows the full transfer matrix: training on language $X$ and testing on language $Y$, for both TP and Morfessor, with no test-language hyperparameter tuning.

\begin{table*}[t]
\centering
\small
\begin{tabular}{l|ccc|ccc|c}
\toprule
 & \multicolumn{3}{c|}{\textbf{TP} (test F1)} & \multicolumn{3}{c|}{\textbf{Morfessor} (test F1)} & \\
\textbf{Train} & AKK & SUX & ELX & AKK & SUX & ELX & \textbf{Train tokens} \\
\midrule
AKK & .971$^\dagger$ & .967 & .992 & .997$^\dagger$ & .981 & .984 & 1.26M \\
SUX & .960 & .972$^\dagger$ & .993 & \textbf{.859} & .995$^\dagger$ & .953 & 146K \\
ELX & .959 & .969 & .989$^\dagger$ & \textbf{.304} & \textbf{.267} & .985$^\dagger$ & 13K \\
\bottomrule
\end{tabular}
\caption{Cross-language word boundary transfer (F1). $^\dagger$ = in-language (held-out CV). Off-diagonal cells = zero-shot transfer with no test-language hyperparameter tuning. Bold = $>$10 F1 point gap from TP. Train tokens are Table~\ref{tab:datasets} word counts; in segmentation-stream units the corpora are 2.44M / 338K / 8.8K (ELX segmentation trains on the Utu-na\v{s}u running text only, \S\ref{sec:data}). 95\% bootstrap confidence intervals (1,000 document resamples) are within $\pm$0.006 across all transfer cells; the cross-language dissociation between TP and Morfessor (65+ F1 points at ELX-as-source) is therefore well outside any reasonable noise band. Visualized in Figure~\ref{fig:transfer_bars}.}
\label{tab:transfer}
\end{table*}

When the training corpus is large (Akkadian at 1.26M tokens), both methods transfer reasonably well: AKK$\rightarrow$SUX yields TP F1 = 0.967 and Morfessor F1 = 0.981; AKK$\rightarrow$ELX yields TP F1 = 0.992 and Morfessor F1 = 0.984. The methods are roughly comparable in this high-resource regime, with Morfessor holding a small edge on AKK as the source. As the training corpus shrinks, however, the methods diverge sharply. Training on Sumerian (146K tokens), TP transfers to Akkadian at F1 = 0.960 while Morfessor drops to F1 = 0.859, a 10-point gap. Training on the small Elamite corpus (13K tokens), TP transfers to Akkadian at F1 = 0.959 while Morfessor collapses to F1 = 0.304; TP transfers to Sumerian at F1 = 0.969 while Morfessor collapses to F1 = 0.267. The gaps in the low-resource transfer setting exceed 65 F1 points.

This dissociation has a clean substrate-decomposition explanation. TP's sufficient statistics (sign-level bigram probabilities) appear to function, across the three corpora we test, as properties of a shared script substrate: within-word sign sequences in Akkadian, Sumerian, and Elamite obey similar local statistical regularities, consistent with their shared use of the cuneiform writing system, even when the underlying vocabulary and morphology differ completely.\footnote{This shared substrate was historically grounded in the cuneiform scribal tradition. Scribal schools (Sumerian \textit{edubba'a}) trained students through repeated copying of sign lists, syllabaries, lexical lists, and model texts that circulated for centuries across Mesopotamia and neighboring cuneiform-writing cultures. These curricular traditions contributed to the standardization and long-term transmission of sign-level orthographic conventions across Sumerian, Akkadian, Hittite, and Elamite scholarly milieus \cite{sjoberg1975,tinney1999,robson2001,veldhuis2004}.} Morfessor's learned lexicon, by contrast, is language-specific: a model trained on Elamite memorizes Elamite words, which appear essentially nowhere in Akkadian or Sumerian text, so cross-language failure reflects the lexicon's \textit{coverage} of the target corpus (to which both source size and source composition contribute; see below) rather than any failure of the algorithm itself. This interpretation is corroborated by the in-language learning curve below (\S\ref{sec:curve}): Morfessor at 10K Akkadian training tokens reaches F1 $>$ 0.95 on a held-out Akkadian test, while Morfessor at 13K Elamite training tokens reaches only F1 = 0.30 on Akkadian test. The 65+ point gap between these two settings, both at comparable training-corpus size, isolates the vocabulary-transfer mechanism from any pure sample-efficiency confound. We release the TP segmenter as \textsc{cunei-seg}, with pre-trained models for all three languages.

\paragraph{Coverage, not language identity, drives the collapse.} A matched-size control proposed during review sharpens the mechanism claim. Training Morfessor on Sumerian subsampled to the Elamite training budget (5 seeds, whole-document subsampling, zero-shot evaluation) separates two contributions:

\begin{table}[t]
\centering
\small
\begin{tabular}{lcc}
\toprule
\textbf{Morfessor source} & $\rightarrow$\textbf{AKK} & $\rightarrow$\textbf{ELX} \\
\midrule
SUX @ 13K stream tokens & .595 $\pm$ .031 & .710 $\pm$ .091 \\
SUX @ 8{,}831 (ELX budget) & .542 $\pm$ .051 & .637 $\pm$ .077 \\
SUX full (Table~\ref{tab:transfer}) & .857 & .953 \\
ELX full (Table~\ref{tab:transfer}) & .304 & --- \\
\bottomrule
\end{tabular}
\caption{Matched-size transfer control. Shrinking the Sumerian source to the Elamite budget costs 32 points ($\rightarrow$AKK), yet at the same budget the Elamite source sits a further 24 points lower: size and composition both matter, and neither alone explains the collapse.}
\label{tab:matched}
\end{table}

Both axes act through a single quantity: the lexicon's coverage of the target. The 13K-token Sumerian lexicon covers 96.9\% of Elamite target tokens (hence SUX$\rightarrow$ELX holds at .710), while the Elamite lexicon covers only 71.7\% of Sumerian. Across a full grid of 14 source-size configurations (5 seeds each, 140 evaluations; Appendix~\ref{app:grid}), coverage of the target predicts Morfessor's transfer F1 at Pearson $r = 0.91$. A causal test completes the argument: training Morfessor on the \textit{full} Sumerian corpus with its lexicon clamped to the Elamite sign inventory collapses transfer to $\rightarrow$AKK $= 0.283$ (unclamped: 0.857), and the mirror clamp on Akkadian gives $\rightarrow$SUX $= 0.285$ (unclamped: 0.981). Holding the language constant and intervening only on inventory coverage reproduces the Table~\ref{tab:transfer} collapse. In-language sample efficiency (\S\ref{sec:curve}) and cross-language transfer are thus two views of the same coverage mechanism, while TP requires no lexicon at all.

\paragraph{What the script-level substrate consists of.} The coverage asymmetry is itself a quantitative trace of the shared scribal tradition sketched in the footnote above: the Sumerian lexicon blankets Elamite because syllabary conventions and high-frequency sign sequences propagated from Sumerian scribal curricula outward, while Elamite's leaner inventory covers only a fraction of Sumerian. The regularities TP exploits are correspondingly graphemic rather than lexical: transitions out of determinatives reliably mark word-initial position across all three languages, and syllabic-structure constraints differ systematically at word edges as against word-internal sequences, so the sign-bigram statistics that signal a boundary in Akkadian remain informative in Sumerian and Elamite even though the words themselves never recur.

\begin{figure}[t]
\centering
\includegraphics[width=\columnwidth]{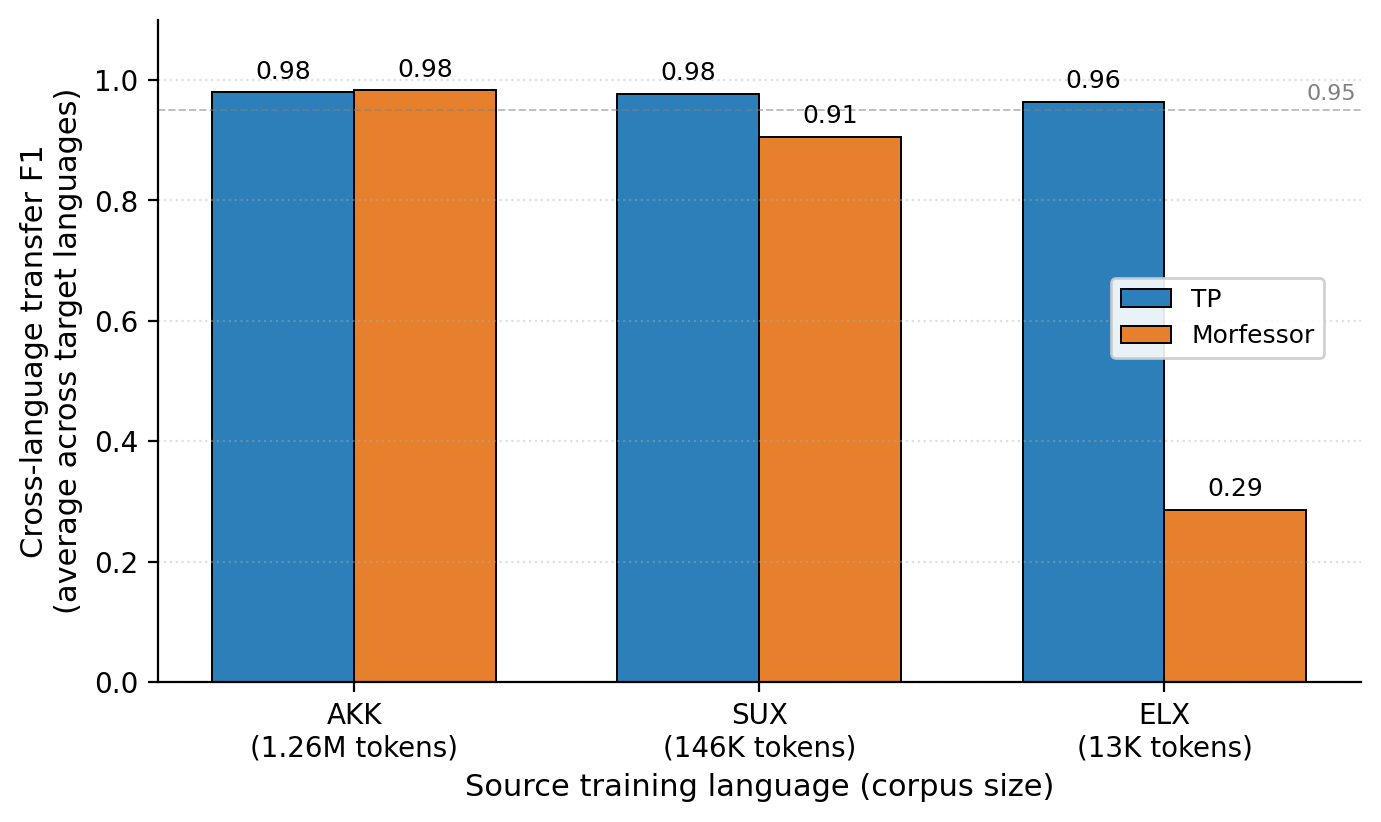}
\caption{Cross-language transfer F1 by source language and method. The dissociation is corpus-size mediated: TP (blue) maintains F1 $>$ 0.95 across all source corpora, while Morfessor (orange) tracks TP at the largest source (AKK, 1.26M tokens), trails by approximately 10 points at the medium source (SUX, 146K tokens), and collapses to below 0.31 at the smallest source (ELX, 13K tokens). The pattern is exactly what would be expected if TP requires only stable bigram statistics (which converge from approximately $10^4$ tokens) while Morfessor requires a lexicon that \textit{covers} the target corpus; coverage, not language identity per se, predicts its transfer (\S\ref{sec:transfer}).}
\label{fig:transfer_bars}
\end{figure}

\subsubsection{Recoverability of the Script-Level Substrate from Minimal Data}
\label{sec:curve}

\begin{figure}[t]
\centering
\includegraphics[width=\columnwidth]{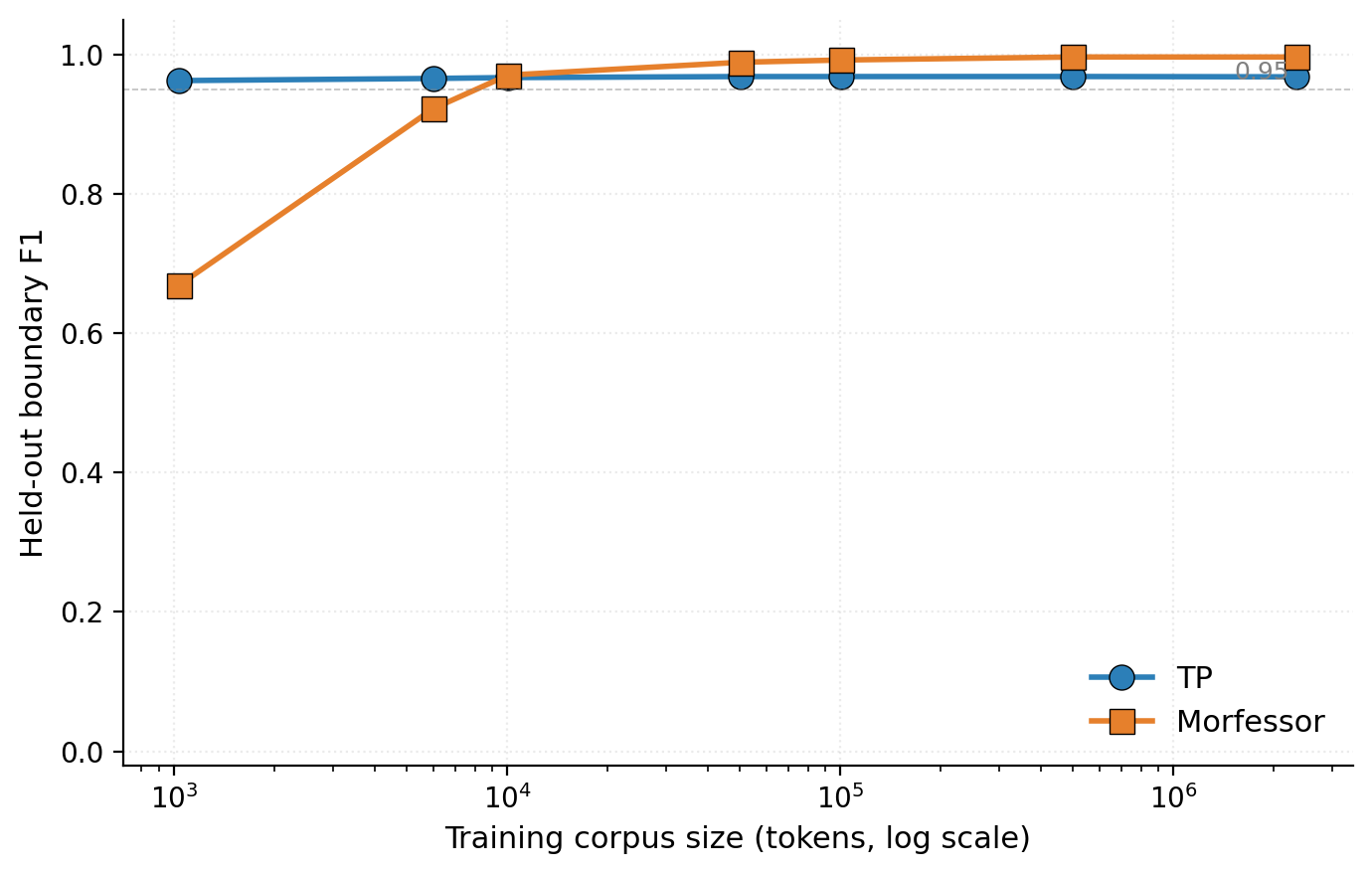}
\caption{In-language learning curve for TP (blue) and Morfessor (orange) on Akkadian, evaluated on a fixed held-out test set. TP is essentially flat at F1 $\approx$ 0.96 across all training-corpus sizes from 1K to 1.26M tokens, confirming that the script-level substrate property is recoverable from minimal data. Morfessor starts at F1 = 0.67 at 1K tokens, climbs to F1 $\approx$ 0.92 at 5K, and reaches TP's level by approximately 10K tokens; above the crossover, Morfessor slightly exceeds TP. The in-language curve isolates property-recoverability (the focus of this experiment) from cross-language vocabulary overlap (the focus of Table~\ref{tab:transfer}); together they account for the full cross-language dissociation.}
\label{fig:learning_curve}
\end{figure}

The cross-language transfer matrix shows that TP's substrate property transfers across languages while Morfessor's does not. A complementary question is how much training data each method requires to learn its respective substrate property in the first place. We run a controlled learning curve on a single language to isolate this question from the cross-language vocabulary-overlap confound. We subsample the Akkadian training corpus to $\{10^3, 5\!\cdot\!10^3, 10^4, 5\!\cdot\!10^4, 10^5, 5\!\cdot\!10^5, 1.26\!\cdot\!10^6\}$ tokens, train both TP and Morfessor at each scale, and evaluate on a fixed held-out Akkadian test set (Figure~\ref{fig:learning_curve}).

TP is near asymptote at the smallest training size we test: F1 = 0.961 at 1K tokens, climbing only marginally to F1 = 0.967 at 1.26M tokens. Morfessor starts at F1 = 0.67 at 1K, reaches F1 = 0.92 at 5K, and matches TP near 10K training tokens; from 50K onward, Morfessor slightly exceeds TP. The crossover at approximately $10^4$ tokens tells us two things. First, TP's sufficient statistics converge from very little data, consistent with the script-level substrate interpretation. Second, Morfessor at 10K in-language tokens reaches F1 $>$ 0.95, while Morfessor at 13K Elamite tokens tested on Akkadian collapses to F1 = 0.30 (Table~\ref{tab:transfer}); the 65-point gap at comparable corpus size isolates the coverage mechanism (\S\ref{sec:transfer}) from sample efficiency. The two experiments together decompose the substrate, in the corpora we study, into a recoverability dimension (TP needs less data) and a transfer dimension (TP's property holds across the three languages we test).

\subsubsection{Hittite: Substrate Decomposition}
\label{sec:hittite}

The cross-language transfer and learning curve results confirm one prediction of the theory: simple statistical methods exploit substrate properties that converge from limited data. A second prediction is that when the substrate itself is violated, both methods should degrade, but through different mechanisms and to different degrees, depending on which aspect of substrate each method depends on. Hittite provides a natural test. Hittite cuneiform inserts \textit{Akkadograms} (Akkadian words read as Hittite) and \textit{Sumerograms} (Sumerian logograms read as Hittite) inline within Hittite syllabic text, often mid-word. This is a within-word substrate violation: it injects a heterogeneous sign sequence between Hittite syllabic signs, breaking the within-word continuity that both methods, in their different ways, rely on.

Trained and evaluated on Hittite documents under the same protocol as the three other languages, TP achieves F1 = 0.619 (best at $\theta = 0.06$) while Morfessor achieves F1 = 0.918, both below the F1 $>$ 0.96 baseline established for Akkadian, Sumerian, and Elamite. The degradation is sharply asymmetric: TP loses 35 F1 points relative to its three-language average, Morfessor 8. We attribute this asymmetry to the structure of Hittite cuneiform, in which 34.7\% of tokens are foreign-script logograms detected via a conservative multi-character uppercase Latin heuristic (representative examples include LUGAL `king' with 420 attestations, NINDA.GUR\textsubscript{4}.RA `thick bread' with 1{,}238, and MUNUS.LUGAL `queen' with 484). These insertions affect the two methods through different mechanisms. TP exploits within-word sign-level statistical continuity: a Hittite syllabic sign abutting a foreign-script logogram produces an unstable local transition probability that does not reliably mark word boundaries. Morfessor exploits lexical-unit recurrence: foreign-script logograms still recur as consistent units within the Hittite corpus, and Morfessor's memorized vocabulary captures them despite the script-mixing. The asymmetric failure identifies the specific aspect of cuneiform substrate that each method depends on, and shows that script-mixing breaks within-word sign-level continuity more severely than lexical-unit recurrence.

\paragraph{Robustness of the logogram rate.} The 34.7\% rate is our quantitative anchor for the substrate-violation mechanism. Varying the detector heuristic leaves every variant in the 25--35\% range (Appendix~\ref{app:hittite}), so the substantive claim, that more than a quarter of Hittite tokens are foreign-script logograms violating within-word continuity, survives under any reasonable detector. Hittite serves throughout as a substrate-violation stress test rather than a fourth evaluation language. An Old Babylonian case study on naturally mixed Akkadian--Sumerographic letters extends the same analysis diachronically and shows both failure modes of the decomposition co-occurring (Appendix~\ref{app:ob}).

\subsection{POS Classification}
\label{sec:pos}

Having compared segmentation methods, we next ask whether the choice of representation (Latin vs.~Unicode) affects downstream classification, using the same script-versus-language probe-pair logic. We compare Latin and Unicode on token-level part-of-speech classification using character n-gram logistic regression. The unified ORACC POS tagset combines grammatical categories with named-entity tags \cite{robson_tinney2019}. Table~\ref{tab:pos} reports macro-F1. Latin outperforms Unicode for Akkadian and Sumerian, while Unicode wins for Elamite, hinting that corpus size mediates the representation advantage. Sumerian's high absolute scores partly reflect ETCSL's curated literary annotation together with its smaller tagset (14 classes versus 24 for Akkadian; Table~\ref{tab:datasets}), which raises macro-F1 chance baselines. The separated grammatical-POS-versus-entity analysis, which reveals a hidden Unicode advantage on grammatical categories that the unified tagset masks, is reported in Appendix~\ref{app:separated}.

\begin{table}[!ht]
\centering
\small
\begin{tabular}{llccc}
\toprule
\textbf{Task} & \textbf{Repr.} & \textbf{AKK} & \textbf{SUX} & \textbf{ELX} \\
\midrule
\multirow{3}{*}{Unified}
 & Latin & .693 & .847 & .636 \\
 & Unicode & .621 & .750 & \textbf{.659}$^{**}$ \\
 & Concat & \textbf{.712}$^{***}$ & \textbf{.870}$^{***}$ & \textbf{.666}$^{***}$ \\
\bottomrule
\end{tabular}
\caption{POS classification (F1-macro, n-gram LR). Significance markers are paired-bootstrap $p$-values comparing Concat against the better of Latin and Unicode within each column: $^{***}p < 0.01$; $^{**}p < 0.02$. Concatenation wins across all three languages. Separated grammatical-vs-entity results in Appendix~\ref{app:separated}.}
\label{tab:pos}
\end{table}

\subsection{Representation Concatenation as a Second Substrate Probe}
\label{sec:concat}

The segmentation results identify two separable substrate properties in the cuneiform corpora we study: a script-level property that transfers across the three languages tested and is recoverable from minimal data, and a language-level property that does not transfer. We now ask whether the same decomposition appears at the representational level. Unicode preserves graphemic distinctions and is largely language-invariant (the same Unicode glyph appears in Akkadian, Sumerian, Elamite, and Hittite text). Latin transliteration encodes interpreted reading values that are language-specific (the same sign A U+12000 transliterates as `a' (water) in Sumerian but enters Akkadian text as the syllabic value /a/, with different conventional surroundings in transliteration). If Unicode and Latin behave like a second probe pair for the same script-versus-language decomposition, we should observe complementary information: Unicode for script-level patterns, Latin for language-specific interpretation, and the best system combining both.

We test this by concatenating Latin and Unicode character n-gram feature vectors and training a single logistic regression classifier on the combined representation. If the representations were redundant, this should not help. The Concat row in Table~\ref{tab:pos} shows the opposite: concatenation wins across all three languages with the unified tagset, and the same pattern holds on the separated grammatical and entity subtasks (Appendix~\ref{app:separated}). Across all 12 task-language comparisons, concatenation wins 12 of 12 with 11 statistically significant at $p < 0.01$ (paired bootstrap); the sole exception is Akkadian entity typing ($p = 0.16$). The complementarity is robust to architecture: a 2-layer character Transformer trained from scratch shows concatenation wins 5 of 6 task-language pairs, combining to 17 of 18 wins across both architectures (Appendix~\ref{app:robustness}). Character-level seq2seq lemmatization is the one task where Latin alone wins across all three languages, since lemma targets are themselves in Latin script (Appendix~\ref{app:lemma}).

The complementarity reproduces the segmentation result at the representational level: Unicode supplies sign-level distinctions (paralleling TP's substrate property), Latin supplies language-specific reading-value interpretation (paralleling Morfessor's). The two probe pairs converge on the same finding.

\section{The \textsc{cunei-tools} Toolkit}

We release \textsc{cunei-tools}, a pip-installable Python package comprising: (1)~\textsc{cunei-seg}, which implements transitional-probability word boundary detection with pre-trained models for all three languages; and (2)~\textsc{cunei-conv}, which provides bidirectional Latin--Unicode conversion using 18,272 sign mappings.

\section{Conclusion}

This paper characterizes the three cuneiform corpora we study as exhibiting separable script-level and language-specific statistical regularities, exposed through two pairs of probes. Within language, transitional probability reaches F1 $>$ 0.97 and Morfessor reaches F1 $>$ 0.99 on Akkadian and Sumerian given sufficient data, surpassing the prior dictionary-based MaxMatch baseline by more than 23 F1 points and constituting, to our knowledge, the first systematic evaluation of dictionary-free segmentation methods on cuneiform. The two dictionary-free methods diverge under stress in informative ways. In cross-language transfer, TP holds F1 $>$ 0.95 across all six language pairs, consistent with sign-level sufficient statistics that function as properties of a shared script substrate across these corpora, while Morfessor's transfer F1 falls to 0.27 from the small Elamite source, not because of sample size (the in-language learning curve shows Morfessor matches TP at approximately $10^4$ tokens) but because its learned lexicon fails to \textit{cover} the target: across a 14-configuration size grid, lexicon coverage of the target predicts Morfessor's transfer at $r = 0.91$, and clamping a large source to a small sign inventory suffices to reproduce the collapse (\S\ref{sec:transfer}). In the Hittite substrate-violation test, the two methods degrade asymmetrically (TP drops 35 points, Morfessor 8), showing that within-word sign-level continuity is more fragile to script-mixing than lexical recurrence. The same decomposition appears at the representational level: Unicode preserves orthographic distinctions, Latin encodes language-specific interpretation, and the two are complementary across 11 of 12 morphosyntactic tasks. The magnitudes (65+ points cross-language; a 27-point Hittite asymmetry) make method choice binary rather than marginal.

In practice: use TP for cross-language segmentation or in-language tasks with fewer than 10K segmented tokens, and Morfessor above that scale; concatenate Latin and Unicode for classification, but use Latin alone for lemmatization, where target-script alignment matters. Supervision does not change this calculus: a character-BiLSTM tagger needs roughly 50K gold-segmented tokens merely to match label-free Morfessor (Appendix~\ref{app:neural}), and offers no label-free path to a new language.

\section*{Limitations}

Our Elamite corpus is small (13K dictionary entries, 85 running-text documents); results on this language should be interpreted cautiously. A 12.7$\times$ expansion of the Elamite running corpus assembled during review (871 documents; Appendix~\ref{app:elx_expanded}) leaves TP's transfer values essentially unchanged and improves Morfessor's, consistent with the coverage account, while reversing the in-language ordering on the noisier administrative register it introduces. Unicode conversion excluded 14.2\% of Elamite entries; these exclusions may skew comparisons against Latin.

Our datasets also ingest damaged material rather than filtering it: damage markers are stripped but the flagged signs are kept, and only fully illegible or unlemmatizable tokens are excluded, so all reported results are measured on damage-inclusive text. The damage-marked share of ingested tokens is 27.4\% for Akkadian (a further 22.9\% of raw rows are excluded up front as unlemmatizable), 8.4\% for the Utu-na\v{s}u Elamite corpus, 21.9\% for the expanded Susa material, and 0.6\% for Sumerian, where ETCSL's composite editions resolve damage editorially. Segmentation performance on damage-heavy text is therefore already reflected in our numbers, but genre and preservation quality co-vary with language in ways we cannot fully disentangle. ETCSL Sumerian consists of literary texts; administrative Sumerian may behave differently. Our Hittite logogram detector is a conservative heuristic (multi-character uppercase Latin); the true logogram rate may be higher than the 25--35\% range we report. Our sample-efficiency learning curve is run on Akkadian only; we expect the qualitative shape (TP plateau before Morfessor) to hold for other languages and other corpus sizes, but verifying this for individual under-documented cuneiform languages requires more annotated data than is currently available. Our substrate-level claims should be read as scoped to the specific corpora and dialect/register combinations we evaluate (ORACC Neo-Assyrian and state archives, ETCSL literary Sumerian, the Elamite Dictionary and Utu-na\v{s}u corpus). Diachronic variation within cuneiform languages (for example, Old Babylonian versus Neo-Assyrian Akkadian) and register variation (literary versus administrative texts) may produce different boundary statistics; whether the substrate property we identify generalizes across these dimensions is an empirical question we leave to future work.

Our evaluation focuses on morphosyntactic tasks. Representation effects on higher-level tasks such as machine translation remain unexplored. Our ByT5 evaluation reveals that pretrained byte-level models cannot effectively exploit cuneiform Unicode without script-specific pretraining. Sign lists have gaps, and some Unicode mappings are approximations. Both Unicode encoding and scholarly transliteration necessarily abstract over historically variable scribal practices.

\section*{Ethics Statement}

All data comes from published scholarly editions of ancient texts. No human subjects were involved. We release all code and data to support reproducibility.

\section*{Use of AI Assistance}

We used AI assistance (a large language model) for prose editing, LaTeX formatting, and code debugging during paper preparation. All research design, experimental methodology, data collection, results, and scientific claims are the authors' own work. AI was not used for research ideation, experimental design, or interpretation of results.

\section*{Acknowledgments}

We thank the anonymous ARR reviewers and area chair, whose suggested controls materially improved the paper's analysis; the ORACC, ETCSL, CEPOAT, and CDLI projects for making their data openly available; and Timo Homburg, together with the creators of the Nuolenna \cite{nuolenna} and Akkademia \cite{gordin2020akkademia} sign lists, whose resources underlie our conversion pipeline.

\bibliography{references}

\appendix
\section{Supplementary Results}

\subsection{Expanded Related Work}
\label{app:related}

This appendix expands the main-body Related Work with detailed citation context. \citet{gutherz2023} developed neural machine translation for Akkadian. \citet{sahala2021} presented a comprehensive computational Assyriology framework. \citet{sahala2022} released BabyLemmatizer, a neural POS tagger and lemmatizer for Akkadian. \citet{sahala2023} developed a neural pipeline for cuneiform POS tagging and lemmatization. For Sumerian, \citet{pageperron2017} explored machine translation and \citet{liu2015} performed unsupervised named entity recognition. \citet{punia2020} developed the first neural MT system for Sumerian transliterations. At the sign level, \citet{dencker2020} applied deep learning for cuneiform sign detection, and \citet{gordin2024} developed OCR systems for cuneiform. \citet{williams2025} applied deep learning to Elamite cuneiform sign classification. \citet{smidt2024} reported accuracy losses of up to 24.7 points with Unicode input on Akkadian POS tagging.

For Arabic NLP, diacritization and romanization effects have been studied extensively \cite{darwish2014}. \citet{li2019} compared character-level, word-level, and pinyin representations for Chinese segmentation. The substrate decomposition framework we propose generalizes these script-versus-language tradeoffs by isolating the underlying mechanism: which method or representation exploits writing-system-level versus language-level statistical structure.

\subsection{Separated POS vs.\ Entity Classification}
\label{app:separated}

The unified ORACC POS tagset conflates two fundamentally different classification problems: grammatical category assignment and named entity recognition. We separate the evaluation into three subtasks: (1)~grammatical POS with all entity tokens excluded, (2)~binary entity detection, and (3)~fine-grained entity typing. This separation uncovers a hidden finding. Table~\ref{tab:pos_separated} shows that Unicode significantly outperforms Latin on Akkadian \textit{grammatical POS} when entities are excluded ($p = 0.02$), a result invisible in the unified tagset. Latin's strong advantage on entity classification ($\Delta = +0.098$, $p < 0.01$) was masking Unicode's grammatical advantage. For Elamite, Unicode wins or ties on all three subtasks.

\begin{table}[!ht]
\centering
\small
\begin{tabular}{llccc}
\toprule
\textbf{Task} & \textbf{Repr.} & \textbf{AKK} & \textbf{SUX} & \textbf{ELX} \\
\midrule
\multirow{3}{*}{Gram.}
 & Latin & .669 & .872 & .615 \\
 & Unicode & .657$^{*}$ & .832 & .634$^{**}$ \\
 & Concat & \textbf{.682}$^{***}$ & \textbf{.888}$^{***}$ & \textbf{.642}$^{***}$ \\
\midrule
\multirow{3}{*}{Ent.\ det.}
 & Latin & .937 & .947 & .873 \\
 & Unicode & .839 & .896 & .876$^{\text{ns}}$ \\
 & Concat & \textbf{.939}$^{***}$ & \textbf{.954}$^{***}$ & \textbf{.882}$^{***}$ \\
\midrule
\multirow{3}{*}{Ent.\ type}
 & Latin & .897 & .957 & .871 \\
 & Unicode & .833 & .923 & \textbf{.881}$^{*}$ \\
 & Concat & \textbf{.910}$^{\text{ns}}$ & \textbf{.963}$^{***}$ & \textbf{.883}$^{***}$ \\
\bottomrule
\end{tabular}
\caption{Separated grammatical POS, entity detection, and entity typing (F1-macro). Significance markers are paired-bootstrap $p$-values comparing Concat against the better of Latin and Unicode within each column: $^{***}p<0.01$; $^{**}p<0.02$; $^{*}p<0.05$. Concatenation wins all 9 comparisons; 8 are significant.}
\label{tab:pos_separated}
\end{table}

\subsection{Lemmatization}
\label{app:lemma}

Character-level LSTM seq2seq models map surface forms to lemmas. Latin wins across all three languages (AKK: 68.8\% vs.\ 64.4\%; SUX: 81.9\% vs.\ 77.5\%; ELX: 42.5\% vs.\ 25.8\% exact match). This is expected, since lemma targets are themselves in Latin script, and confirms that representation advantages are task-dependent.

\subsection{Architecture Robustness}
\label{app:robustness}

To test whether the complementarity effect is an artifact of linear models, we train a 2-layer character Transformer encoder (128-dim, 4-head, trained from scratch) on the same tasks. For the concatenated condition, both representations are fed as a single sequence separated by a delimiter.

\begin{table}[!ht]
\centering
\small
\begin{tabular}{llccc}
\toprule
\textbf{Task} & \textbf{Lang} & \textbf{Best} & \textbf{Cat.} & $\Delta$ \\
\midrule
\multirow{3}{*}{Uni.}
  & AKK & .373 & \textbf{.412} & +.039 \\
  & SUX & .428 & \textbf{.451} & +.023 \\
  & ELX & .471 & \textbf{.489} & +.018 \\
\midrule
\multirow{3}{*}{Gram.}
  & AKK & \textbf{.516} & .513 & $-$.003 \\
  & SUX & .630 & \textbf{.654} & +.024 \\
  & ELX & .465 & \textbf{.469} & +.004 \\
\bottomrule
\end{tabular}
\caption{Character Transformer: best single representation vs.\ concatenation. Concat wins 5/6 Transformer comparisons; combined with logistic regression (Table~\ref{tab:pos}): concat wins 17/18 across both architectures.}
\label{tab:transformer}
\end{table}

The Transformer's Unicode preference parallels Chinese NLP findings \cite{glyce2019,chinesebert2021}, where self-attention benefits disproportionately from logographic glyph information. ByT5-small \cite{xue2022}, a byte-level pretrained model, strongly favors Latin over Unicode (e.g., AKK unified: .571 vs.\ .334). This result is consistent with the substrate-decomposition framework rather than against it: ByT5's pretraining corpus is overwhelmingly Latin-script and contains essentially no cuneiform Unicode, so the model's representational machinery is tuned to recover language-specific lexical patterns from Latin-encoded text. Unicode cuneiform offers the same script-level substrate to ByT5 that it offers to TP, but ByT5 cannot exploit it because its prior knowledge does not include the relevant sign-level statistical regularities. The from-scratch character Transformer above, by contrast, can learn the script-level substrate during training and shows the predicted Unicode advantage. ByT5 therefore illustrates that the script-level substrate is exploitable only by models whose representational capacity is aligned with it, either by training from scratch (the character Transformer) or by carrying no prior bias (TP's parameter-free statistics).

\subsection{Feature Importance: What Does Each Representation Encode?}
\label{app:features}

\begin{figure}[!ht]
\centering
\includegraphics[width=\columnwidth]{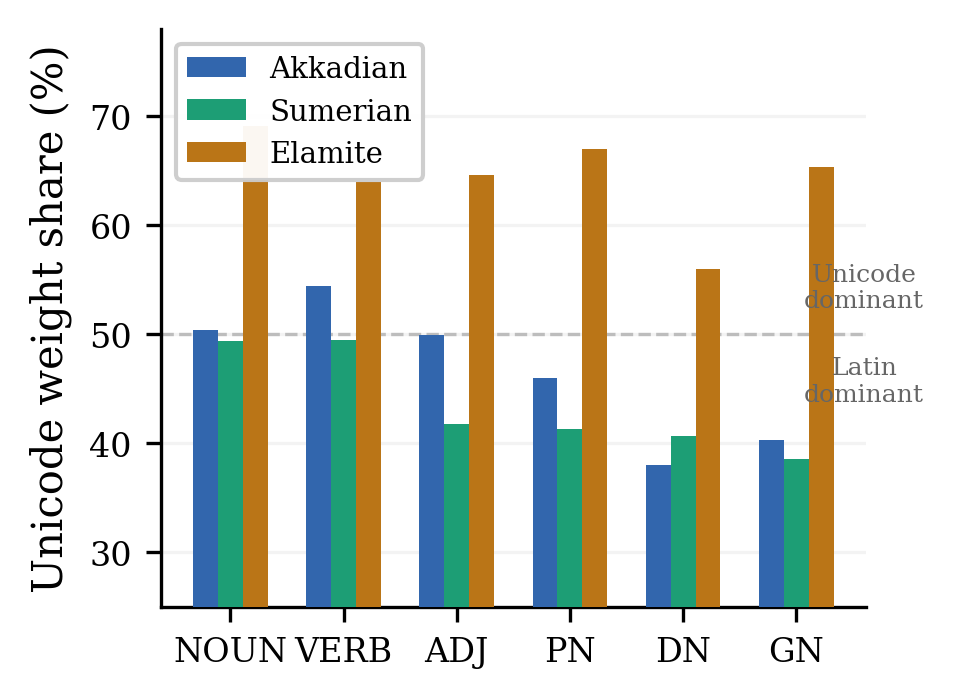}
\caption{Unicode feature weight share by POS category. Elamite is Unicode-dominant; Akkadian and Sumerian show balanced contributions for open-class words but Latin dominance for entities and closed-class items.}
\label{fig:features}
\end{figure}

To understand why concatenation helps, we examine the logistic regression model trained on concatenated features. For each POS class, we compute what fraction of total absolute weight falls on Unicode-side versus Latin-side features (Figure~\ref{fig:features}).

For Akkadian and Sumerian, open-class content words (NOUN, VERB) draw roughly equally from both representations (49--54\% Unicode share), while closed-class categories (DET: 29\%, CONJ: 34\%) and named entities (DN: 38--41\%, GN: 39--40\%) lean heavily toward Latin. (Per-language class inventories and entity ratios for these categories are those of Table~\ref{tab:datasets}.) For Elamite, Unicode features dominate \textit{every} category (56--69\%). Nouns and verbs in cuneiform tend to be compositionally rich (multiple signs encoding stem, tense, person), where Unicode captures sign-level compositional patterns naturally. Closed-class functional words are typically short (one or two signs), where the Latin reading value provides more discriminative power than sign identity alone.

\subsection{Error Analysis}
\label{app:errors}

For Akkadian, 18.4\% of tokens are correctly classified by one representation but not the other (Table~\ref{tab:errors}). The effect is mediated by sign count: for Sumerian, Latin wins on single-sign words by a 13:1 ratio, where polyphony makes a single Unicode glyph ambiguous. As sign count increases, Unicode's compositional advantage grows, reaching parity at 4--5 signs.

\begin{table}[!ht]
\centering
\small
\begin{tabular}{lrrrr}
\toprule
 & \textbf{Both} & \textbf{Both} & \textbf{Latin} & \textbf{Uni.} \\
 & \textbf{correct} & \textbf{wrong} & \textbf{only} & \textbf{only} \\
\midrule
AKK & 63.1\% & 18.5\% & 12.4\% & 6.0\% \\
SUX & 85.2\% & 5.2\% & 7.5\% & 2.1\% \\
ELX & 66.9\% & 20.2\% & 5.5\% & 7.3\% \\
\bottomrule
\end{tabular}
\caption{Error outcome distribution. ``Latin only'' and ``Unicode only'' columns represent the complementary pool (9.6--18.4\%).}
\label{tab:errors}
\end{table}

The confusion patterns are linguistically revealing. Unicode's most common Akkadian error is misclassifying nouns as divine names (DN): it recognizes the DINGIR determinative sign but cannot distinguish its semantic from its logographic use. Latin resolves this via contextual reading. The complementary pool is largest for rare tokens (Table~\ref{tab:rare_app}): in Akkadian, 28.9\% of rare tokens are correctly classified by only one representation, compared to 19.1\% for frequent tokens.

\begin{table}[!ht]
\centering
\small
\begin{tabular}{lcc}
\toprule
 & \textbf{Rare ($\leq$26)} & \textbf{Frequent ($>$26)} \\
\midrule
AKK & 28.9\% & 19.1\% \\
SUX & 16.3\% & 8.5\% \\
\bottomrule
\end{tabular}
\caption{Percentage of tokens correctly classified by exactly one representation, split by token frequency.}
\label{tab:rare_app}
\end{table}

\subsection{Scale Effects and Embedding Space}
\label{app:embedding}

Unicode's individual advantage decreases with corpus size: Elamite favors Unicode on all tasks while larger corpora generally favor Latin. Yet the concatenation benefit is consistent across all scales, suggesting the complementarity is fundamental rather than a small-data artifact. FastText embeddings reveal a complementary dissociation: Unicode achieves higher morpheme coherence across all three languages (AKK: .796 vs.\ .764; SUX: .839 vs.\ .791; ELX: .644 vs.\ .639), while Latin produces better POS-based silhouette scores and higher POS agreement in nearest-neighbor graphs (AKK: .645 vs.\ .512; SUX: .741 vs.\ .638). Unicode organizes by \textit{form}, Latin by \textit{function}.

\subsection{Embedding Space Metrics}

Table~\ref{tab:embeddings_app} provides full embedding space metrics.

\begin{table}[!ht]
\centering
\small
\begin{tabular}{llccc}
\toprule
\textbf{Metric} & \textbf{Repr.} & \textbf{AKK} & \textbf{SUX} & \textbf{ELX} \\
\midrule
\multirow{2}{*}{Silhouette}
 & Latin & $-$.100 & \textbf{$-$.071} & \textbf{$-$.037} \\
 & Unicode & \textbf{$-$.093} & $-$.147 & $-$.063 \\
\midrule
\multirow{2}{*}{Morph.~coh.}
 & Latin & .764 & .791 & .639 \\
 & Unicode & \textbf{.796} & \textbf{.839} & \textbf{.644} \\
\midrule
\multirow{2}{*}{Graph agr.}
 & Latin & \textbf{.645} & \textbf{.741} & \textbf{.460} \\
 & Unicode & .512 & .638 & .452 \\
\bottomrule
\end{tabular}
\caption{Embedding metrics (cosine similarity).}
\label{tab:embeddings_app}
\end{table}

\subsection{ByT5 Detailed Results}

Table~\ref{tab:byt5_app} provides ByT5-small POS classification results across both task configurations. Latin strongly outperforms Unicode in every cell, consistent with ByT5's pretraining bias toward Latin-script text. Concatenating Latin and Unicode does not improve over Latin alone in any cell, in contrast to the from-scratch character Transformer (Table~\ref{tab:transformer}). This confirms that pretrained byte-level models cannot exploit the cuneiform script substrate without script-specific pretraining: ByT5's representational machinery has been tuned on Latin-script data and cannot recover sign-level statistical regularities from Unicode cuneiform, even when the relevant graphemic information is present in the input.

\begin{table}[!ht]
\centering
\small
\begin{tabular}{llccc}
\toprule
\textbf{Task} & \textbf{Repr.} & \textbf{AKK} & \textbf{SUX} & \textbf{ELX} \\
\midrule
\multirow{3}{*}{Unified}
 & Latin & .571 & .671 & .564 \\
 & Unicode & .334 & .422 & .543 \\
 & Concat & .548 & .650 & .571 \\
\midrule
\multirow{3}{*}{Gram.}
 & Latin & .542 & .728 & .529 \\
 & Unicode & .401 & .508 & .516 \\
 & Concat & .529 & .705 & .534 \\
\bottomrule
\end{tabular}
\caption{ByT5-small POS classification (F1-macro). Latin strongly outperforms Unicode, consistent with Latin-script pretraining bias.}
\label{tab:byt5_app}
\end{table}

\subsection{TP Threshold Sweep}

Table~\ref{tab:sweep_app} reports TP word boundary F1 across a sweep of decision thresholds for each language. The F1 surface is relatively flat for Akkadian and Sumerian across a wide range (within 0.01 of the maximum for $\theta \in [0.30, 0.90]$ on both languages), while Elamite shows a steeper dependence on threshold (F1 climbs from 0.591 at $\theta = 0.20$ to 0.989 at $\theta = 0.95$). The differing optima reflect language-specific bigram distributions: Elamite's small corpus produces sparser bigram counts, requiring a more permissive threshold to detect word boundaries. The flatness near the optimum on Akkadian and Sumerian explains why source-tuned thresholds transfer reasonably well across these two languages despite slight per-language differences, and why the cross-language transfer matrix (Table~\ref{tab:transfer}) is robust to the precise threshold choice.

\begin{table}[!ht]
\centering
\small
\begin{tabular}{lccc}
\toprule
$\theta$ & \textbf{AKK} & \textbf{SUX} & \textbf{ELX} \\
\midrule
0.20 & .963 & .957 & .591 \\
0.30 & .967 & .966 & .685 \\
0.50 & .968 & .971 & .897 \\
0.70 & .971 & .971 & .954 \\
0.90 & .961 & .956 & .984 \\
0.95 & .949 & .943 & .989 \\
\bottomrule
\end{tabular}
\caption{TP word boundary F1 at selected thresholds.}
\label{tab:sweep_app}
\end{table}

\subsection{Hyperparameter Robustness: Morfessor Corpusweight}
\label{app:cw}

Morfessor's single hyperparameter, corpusweight, controls the lexicon-size/likelihood trade-off. Table~\ref{tab:cw_inlang} sweeps it in-language: F1 is stable for corpusweight $\leq 1$ but collapses catastrophically above it, correcting an overly broad stability statement in the submitted version of \S\ref{sec:segmentation}. Source-side selection (in-language CV on the \textit{source}, the only tuning admissible for zero-shot transfer) picks corpusweight $= 0.5$ in all three languages, decisively so only for Elamite ($+0.014$ over 1.0, beyond fold noise; Akkadian $+0.0001$ and Sumerian $+0.0014$ are within it). Table~\ref{tab:cw_transfer} re-runs the full transfer matrix at the selected value: Morfessor improves everywhere, most dramatically from the Elamite source ($\rightarrow$AKK $0.304 \rightarrow 0.765$), yet remains 19--26 F1 points below untuned TP from that source. Hyperparameter tuning therefore softens but does not overturn the dissociation, and the coverage mechanism of \S\ref{sec:transfer} is unchanged: the clamped-inventory intervention at corpusweight 0.5 still collapses transfer ($\rightarrow$AKK $0.732$ from clamped full Sumerian vs.~$0.977$ unclamped).

\begin{table}[!ht]
\centering
\small
\begin{tabular}{lccc}
\toprule
\textbf{corpusweight} & \textbf{ELX} & \textbf{SUX} & \textbf{AKK} \\
\midrule
0.25 & .998 & .994 & .996 \\
0.5  & \textbf{.999} & \textbf{.994} & \textbf{.997} \\
1.0 (paper) & .985 & .993 & .996 \\
2.0 & .063 & .076 & .122 \\
4.0 & .057 & .065 & .065 \\
\bottomrule
\end{tabular}
\caption{In-language 5-fold CV F1 by corpusweight. Bold = source-side selected value.}
\label{tab:cw_inlang}
\end{table}

\begin{table}[!ht]
\centering
\small
\begin{tabular}{llccc}
\toprule
\textbf{Src} & \textbf{Tgt} & \textbf{cw=1.0} & \textbf{cw=0.5} & \textbf{TP} \\
\midrule
AKK & SUX & .981 & .982 & .967 \\
AKK & ELX & .984 & .997 & .992 \\
SUX & AKK & .857 & .977 & .960 \\
SUX & ELX & .953 & .993 & .993 \\
ELX & AKK & .304 & .765 & \textbf{.959} \\
ELX & SUX & .267 & .713 & \textbf{.969} \\
\bottomrule
\end{tabular}
\caption{Zero-shot Morfessor transfer at the published default vs.~the source-selected corpusweight, against untuned TP. Bold = TP's advantage from the smallest source survives tuning.}
\label{tab:cw_transfer}
\end{table}

\subsection{Source-Size Transfer Grid}
\label{app:grid}

Table~\ref{tab:grid} reports the full grid behind the coverage analysis of \S\ref{sec:transfer}: 14 source-size configurations $\times$ 5 seeds (140 evaluations; document-level subsampling; corpusweight 1.0). Across the grid, the Morfessor lexicon's coverage of the target corpus predicts transfer F1 at Pearson $r = 0.91$; transfer swings by more than 60 points as size and composition vary, while TP's published values (0.959--0.995) hold from every source.

\begin{table}[!ht]
\centering
\small
\begin{tabular}{llccc}
\toprule
\textbf{Source} & \textbf{Size} & $\rightarrow$\textbf{AKK} & $\rightarrow$\textbf{SUX} & $\rightarrow$\textbf{ELX} \\
\midrule
AKK & 5K & --- & .615 & .826 \\
AKK & 13K & --- & .743 & .955 \\
AKK & 25K & --- & .826 & .973 \\
AKK & full & --- & .981 & .984 \\
SUX & 5K & .482 & --- & .554 \\
SUX & 13K & .595 & --- & .710 \\
SUX & 25K & .662 & --- & .757 \\
SUX & full & .858 & --- & .953 \\
ELX-exp & 5K & .517 & .414 & --- \\
ELX-exp & 13K & .587 & .478 & --- \\
ELX-exp & 25K & .651 & .527 & --- \\
ELX-exp & full & .783 & .712 & --- \\
ELX-orig & 5K & .273 & .247 & --- \\
ELX-orig & full & .304 & .267 & --- \\
\bottomrule
\end{tabular}
\caption{Morfessor zero-shot transfer across the source-size grid (mean over 5 seeds; seed std $\leq .091$, typically $\leq .03$). ELX-exp = the 871-document expanded corpus (Appendix~\ref{app:elx_expanded}).}
\label{tab:grid}
\end{table}

\subsection{Elamite Training-Source Ablation}
\label{app:elx_ablation}

The published Elamite segmentation results train on the Utu-na\v{s}u running text alone; the Elamite Dictionary enters only classification (\S\ref{sec:data}). Table~\ref{tab:elx_ablation} verifies with identical held-out running-text folds that the dictionary does no hidden segmentation work: TP barely moves across training sources, and for Morfessor, isolated citation forms provide a \textit{worse} lexicon than one-tenth the volume of running text, the same coverage effect as \S\ref{sec:transfer}.

\begin{table}[!ht]
\centering
\small
\begin{tabular}{lccr}
\toprule
\textbf{Training source} & \textbf{TP} & \textbf{Morf.} & \textbf{Tokens} \\
\midrule
Dictionary only & .994 & .727 & 68,251 \\
Utu-na\v{s}u only (paper) & .989 & .985 & 6,759 \\
Combined & .995 & .991 & 75,010 \\
\bottomrule
\end{tabular}
\caption{Elamite segmentation by training source, evaluated on identical held-out running-text folds.}
\label{tab:elx_ablation}
\end{table}

\subsection{Expanded Elamite Corpus}
\label{app:elx_expanded}

During review we expanded the Elamite running corpus from 84 to 871 documents (8.8K to 111{,}733 segmentation-stream tokens, 12.7$\times$) via 787 collated Susa documents from the CEPOAT corpus, using the identical conversion pipeline (90.2\% token conversion). Re-running segmentation: in-language TP falls to $0.923 \pm .007$ (the original-corpus value re-runs unchanged at 0.989) while Morfessor rises to $0.993 \pm .000$, reversing the in-language ordering ($\Delta$F1 $= -.069$, $p < .001$). The reversal is a corpus-composition effect: the Susa material introduces numerals, residual unconverted logograms ($\approx$10\% of tokens), and an administrative register that disrupt TP's local transition statistics, while Morfessor's lexicon absorbs them. This sharpens the practical recommendation of the Conclusion: TP for low-resource settings with cleaner data, Morfessor once corpus size and register heterogeneity grow. Transfer from the expanded source improves for Morfessor ($\rightarrow$AKK $.783$, $\rightarrow$SUX $.712$; Table~\ref{tab:grid}) yet remains 18--25 points below tuned TP, and TP's own transfer is essentially unchanged ($.963$/$.968$ target-tuned; $.924$/$.918$ zero-shot at $\theta = 0.45$).

\subsection{Supervised Neural Baseline}
\label{app:neural}

A character-BiLSTM boundary tagger (the same character-BiLSTM family as Appendix~\ref{app:robustness}) trained on $N$ gold-segmented Akkadian tokens and evaluated on a fixed 500-document held-out set, against TP and Morfessor trained on the identical subsamples:

\begin{table}[!ht]
\centering
\small
\begin{tabular}{lccc}
\toprule
\textbf{Tokens} & \textbf{BiLSTM (sup.)} & \textbf{TP} & \textbf{Morf.} \\
\midrule
1,000 & .957 & .961 & .637 \\
5,000 & .957 & .964 & .908 \\
13,000 & .962 & .968 & .963 \\
50,000 & .994 & .968 & .991 \\
\bottomrule
\end{tabular}
\caption{Supervised vs.~unsupervised boundary detection on identical training subsamples (seed std $\leq .063$; $\leq .01$ except Morfessor at 1K). For calibration, the all-boundaries trivial baseline reaches F1 $= 0.961$ on this boundary-dense test set; the substantive comparisons are therefore precision-driven and lie above that floor.}
\label{tab:bilstm}
\end{table}

Below roughly 50K labeled tokens the supervised tagger does not outperform the unsupervised methods; at 50K it essentially matches what Morfessor learns from the same text without labels (.994 vs.~.991), while presupposing gold segmentation, precisely the resource unavailable for under-documented cuneiform languages.

\subsection{Old Babylonian Case Study}
\label{app:ob}

We test the substrate decomposition diachronically on an Old Babylonian corpus of 132 CDLI documents (13{,}614 words; letters and administrative texts, including the Iltani archive) whose transliteration mixes syllabic Akkadian with Sumerographic writing word by word. Writing mode is labeled per token from the ATF edition markup itself: tokens inside underscore spans are Sumerographic, giving ground-truth labels for 50.3\% of tokens. (An orthographic detector of the kind our Hittite analysis uses reaches only P $= 0.74$ / R $= 0.65$ against this ATF truth, underscoring the value of native markup over heuristics.) Illegible tokens are excluded, and Unicode conversion (91.4\% of tokens; unconverted words split the stream into 1{,}246 analysis runs totaling 11{,}589 tokens) uses the paper's sign mappings. Both methods are trained on the full Neo-Assyrian Akkadian corpus and applied zero-shot, with TP's threshold target-tuned as in the tuned cells of Table~\ref{tab:transfer}.

The register is an extreme version of the Hittite condition: 51\% of word boundaries are writing-mode switches, and words average under three signs. Both methods collapse toward the all-boundaries trivial baseline (F1 $= 0.540$ on this boundary-dense stream): TP reaches F1 $= 0.567$ at its best threshold, and Morfessor over-segments nearly every position (F1 $= 0.551$; within-word false-boundary rate $0.92$--$0.96$). Diagnostics attribute the two collapses to the two different mechanisms the paper identifies. Morfessor's failure is coverage: only 35.4\% of OB tokens appear in the Neo-Assyrian lexicon, exactly the regime of \S\ref{sec:transfer}. TP's failure is not sign novelty (100\% of OB signs and 98.6\% of sign bigrams are attested in training) but transition instability under dense mode-switching, and its residual behavior remains ordered as the substrate account predicts: boundary recall is highest at switch points (0.865) and lowest inside syllabic text (0.709), while false boundaries are markedly rarer inside Sumerographic spans (0.510 vs.~0.659 per within-word position), whose logographic sequences recur as lexicalized units with stable internal transitions. The case study thus shows the framework's two failure modes, lexicon under-coverage and within-word continuity violation, co-occurring in one naturally occurring register, and cautions that segmentation statistics trained on one dialect and register of Akkadian do not transfer to another unchanged (cf.~Limitations).

\subsection{Hittite Experiment Details}
\label{app:hittite}

Hittite corpus \cite{yavasan2024hittite}: 170,496 tokens, 7,099 texts. Unicode conversion rate: 98.6\%. Documents used for segmentation: 5,896 (filtered to those with $\geq 2$ clean Unicode words).

In-language 5-fold CV (held-out, matching the same fold splits across both methods):
\begin{itemize}
\item TP: best F1 = 0.619 at $\theta = 0.06$.
\item Morfessor: F1 = 0.918 ($\pm 0.004$), precision = 0.866, recall = 0.977.
\end{itemize}

Logogram detection: 34.7\% of tokens flagged by the default heuristic (multi-character uppercase Latin after stripping determinative wrappers). Robustness variants: 31.2\% (require $\geq 3$ uppercase letters), 28.4\% (uppercase-only, no mixed case), 26.1\% (exclude period-containing forms). Top detected logograms: NINDA.GUR\textsubscript{4}.RA (1{,}238), 1-\v SU (1{,}225), LUGAL-u\v s (1{,}122), GAL (1{,}101), LUGAL-i (857), KI.MIN (592), S\`IR-RU (572), MUNUS.LUGAL (484), GE\v sTIN (397), and others.

\end{document}